\documentclass{article}

\usepackage{PRIMEarxiv}
\usepackage{multirow}
\usepackage[utf8]{inputenc} 
\usepackage[T1]{fontenc}    
\usepackage{hyperref}       
\usepackage{url}            
\usepackage{booktabs}       
\usepackage{amsfonts}       
\usepackage{nicefrac}       

\usepackage{graphicx}
\usepackage{color}
\usepackage{tabularray}

\usepackage{microtype}      
\usepackage{lipsum}
\usepackage{fancyhdr}       
\usepackage{graphicx}       
\graphicspath{{media/}}     

\newcommand{\BibTeX}{B\kern-.05em{\sc i\kern-.025em b}\kern-.08em\TeX}
\newcommand{\AS}{\mathcal{F}}
\newcommand{\A}{\mathcal{A}}
\renewcommand{\S}{\mathcal{S}}
\newcommand{\C}{\mathcal{C}}

\newcommand{\D}{\mathcal{D}}
  
\title{Assisted Debate Builder with Large Language Models
}

\author{
Elliot Faugier\\
Univ Lyon, UCBL\\
Villeurbanne, France\\
  \texttt{elliotfaugier@gmail.com} \\
    \And
  Frédéric Armetta, Angela Bonifati, Bruno Yun \\
  Univ Lyon, UCBL, CNRS, INSA Lyon, LIRIS,UMR5205, F-69622 \\
   Villeurbanne, France\\
  \texttt{\{frederic.armetta,angela.bonifati, bruno.yun\}@univ-lyon1.fr} \\
}

\begin{document}
\maketitle

\begin{abstract}
We introduce ADBL2, an assisted debate builder tool. 
It is based on the capability of large language models to generalise and perform relation-based argument mining in a wide-variety of domains.
It is the first open-source tool that leverages relation-based mining for (1) the verification of pre-established relations in a debate and (2) the assisted creation of new arguments by means of large language models.
ADBL2 is highly modular and can work with any open-source large language models that are used as plugins.
As a by-product, we also provide the first fine-tuned Mistral-7B large language model for relation-based argument mining, usable by ADBL2, which outperforms existing approaches for this task with an overall F1-score of 90.59\% across all domains.

\end{abstract}


\section{Introduction}

In recent years, there has been a lot of research in artificial intelligence, focusing on leveraging argumentation theory for non-monotonic reasoning \cite{DBLP:conf/sum/CroitoruV13,DBLP:phd/hal/Yun19}. 
Starting with Dung's seminal work \cite{DBLP:journals/ai/Dung95}, many researchers have considered abstract argumentation frameworks, composed of a set of arguments and a binary attack relation between them, and created many semantics for tasks such as computing accepted sets of arguments \cite{DBLP:journals/ker/BaroniCG11, DBLP:conf/comma/Caminada06} or rank arguments \cite{DBLP:conf/sum/AmgoudB13a,DBLP:conf/aaai/BonzonDKM16,DBLP:conf/comma/YunVCB18}.

 This abstract argumentation framework was extended with many features such as supports \cite{DBLP:conf/nmr/AmgoudCL04,DBLP:conf/ecsqaru/CayrolL05,DBLP:conf/sgai/HimeurYBC21}, sets of attacking arguments \cite{DBLP:conf/argmas/NielsenP06,DBLP:conf/aaai/YunVC20}, or probabilities \cite{DBLP:journals/jair/HunterT17} among others. 
 However, one important question that remained was: ``\textit{Where do argumentation frameworks come from in real-life settings?}''.

While there are some pieces of evidence that the fundamental aspects of abstract argumentation frameworks have links with human reasoning \cite{DBLP:conf/jelia/CramerG19,DBLP:conf/atal/VesicYT22}, humans debates or natural language texts are not always written as arguments and the relation between arguments is not always clear, even for experts \cite{DBLP:conf/ijcai/CramerG18}.
The question of the origin of argumentation frameworks is crucial to facilitate the application of argumentation theory semantics in real-world contexts. 

Some online debate platforms like Kialo\footnote{\url{https://www.kialo.com/}}, Debategraph\footnote{\url{https://debategraph.org/}}, Rationale\footnote{\url{https://www.rationaleonline.com/}}, or Argüman\footnote{\url{https://arguman.org/}} allow users to formalise (individually or collaboratively) debates into arguments and attacks/supports. While this constitute a possible source of argumentation frameworks, users are not assisted in the creation of arguments, leading to redundancies, poorly phrased arguments or wrongly classified relations.
We argue that an automatic assistant is essential to help users elicit high quality argumentation frameworks. 
Moreover, this automatic assistant would need to be highly adaptable to a variety of debate domains, thus motivating the need for large language models (LLMs).

In this paper, our contributions are as follows:

\begin{itemize}
    \item ADBL2, an assisted debate builder tool. 
It leverages the capability of large language models to generalise and perform relation-based argument mining (RBAM) in a wide-variety of domains.
While RBAM has been used for several tasks \cite{DBLP:conf/naacl/CarstensT15,DBLP:conf/lrec/KonatLPBR16}, ADBL2 is the first open-source tool that imports debates from Kialo and leverages RBAM for (2) the verification of existing relations in a debate, and (3) assist users in the creation of new arguments.

\item An open-source and fine-tuned Mistral-7B LLM for the task of relation-based argument mining, embedded in ADBL2, which outperforms existing approaches in multiple domains.
\end{itemize}

This demonstration paper is structured as follows. In Section \ref{sec:llmRBAM}, we motivate the use of fine-tuned LLMs for the RBAM task. In Section \ref{sec:tool}, we introduce the architecture and use-cases of ADBL2. In Section \ref{sec:fine-tined}, we explain the data collection, fine-tuning, and evaluation of our LLM. Finally, we conclude and discuss future work in Section \ref{sec:future_work}.

The demo video is available at: \url{https://youtu.be/KMzqKJlH9lE}.

\section{LLMs for Relation-based Argument Mining}
\label{sec:llmRBAM}

Relation-based argument mining is a fundamental task in argument mining and is essential to support online debates and obtain high-quality argumentation frameworks \cite{DBLP:journals/coling/LawrenceR19}.
It consists in the automatic identification of argumentative relations, aiming at determining how
different texts are related within the argumentative discourse.
While RBAM can take many forms, we will focus on the binary version in this paper, i.e., classifying relations as supports or attacks.
For example, given the following three argumentative texts from Kialo. 
 $a_1 = $ ``It is important for sporting bodies to level the playing field among atheletes'', $a_2 = $ ``The knowledge that they will never beat a competitor like Caster Semenya can damage the athlete's mental health'', and  $a_3=$ ``By trying to weed out extraordinary sportswomen to cater for the majority, the sporting community could lose extremely talented atheletes''.
One can infer that $a_2$ supports $a_1$ as it illustrates the potential mental health concerns of not leveling the playing field in sports while $a_3$ attacks $a_1$ by suggesting that leveling the playing field could lead to unintended consequences (i.e., losing exceptionally talented athletes), thus weakening it.
Here, contextual information about individuals (e.g., the identity or characteristics of Caster Semenya) or events (e.g., the breakdown of athlete Lynsey Sharp during the Rio's Olympic 800m final) are important for the prediction.

While there are some small transformer-based models (e.g., BERT-based models) that can perform relatively well on specific datasets by identifying language patterns and learning good latent representation of concepts, they are usually limited to specific domains \cite{DBLP:journals/artmed/MayerMCV21} and fail to generalise across multiple dataset \cite{DBLP:journals/expert/Ruiz-DolzABG21}.
This generalisation capability is essential if one wants to have a single backbone model for a debate assistant tool.

The recent work of Gorur et al. \cite{DBLP:journals/corr/abs-2402-11243} explores the usage of two types of open-source LLMs (Meta AI's Llama-2 models \cite{touvron2023llama} and Mistral AI's models \cite{jiang2023mistral}) for RBAM on ten datasets. They showed that LLMs equipped with few-shot examples (2 pairs of fixed arguments) outperform the RoBERTa baseline. However, while the larger models (70B parameters) had better performances, they also had slower inference time and greater GPU requirements. 
In this paper, we will explore whether fine-tuning smaller LLMs for RBAM can yield similar or better performances.

\section{The ADBL2 Tool}
\label{sec:tool}

ADBL2 is an online tool aiming to ease debate tree construction leveraging LLMs and prompt techniques to help the user formulating arguments which can be unclear.
The source-code of the tool is available at: \url{https://github.com/4mbroise/ADBL2}.

\begin{figure*}[!h]
    \centering
    \includegraphics[width=16cm]{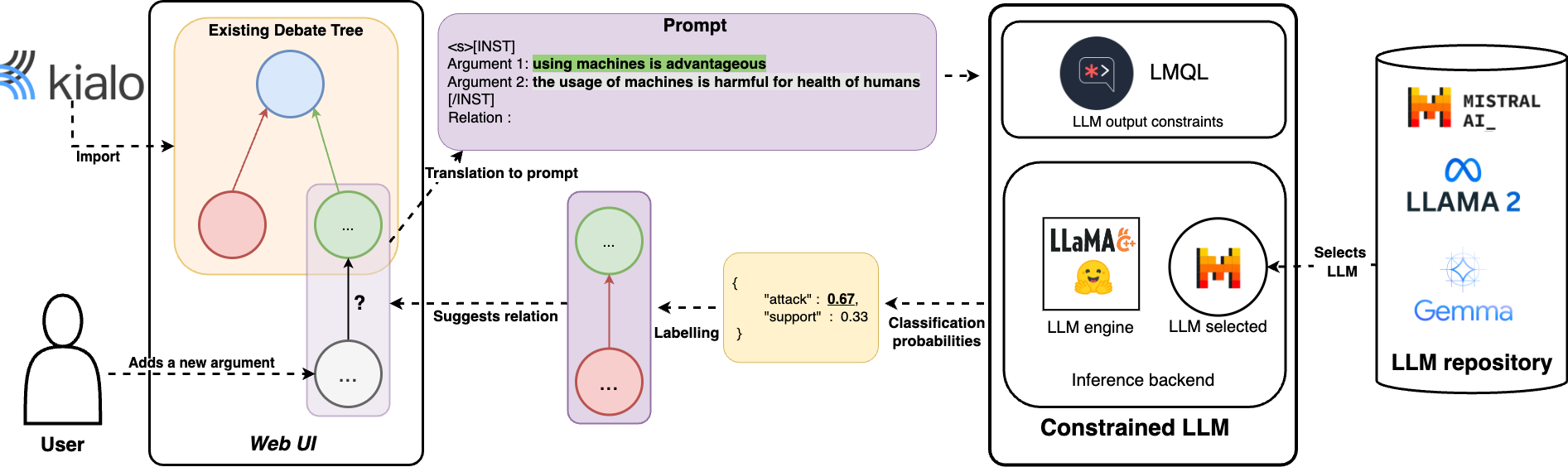}
    \caption{Representation of the architecture of the ADBL2 tool.}
    \label{fig:architecture}
\end{figure*}

ADBL2 allows users to verify existing relations and assist users in the creation of new arguments by relying on its underlying RBAM model.
For example, in the unfolded scenario when one wants to edit an existing argument which is connected to other arguments, it is essential to verify that the existing relations remain the same or to modify them accordingly. In an other scenario where a user wants to add a new argument to a parent argument, the classification probability displayed to the user can help them to modify and refine their textual arguments to achieve the desired effect.

The architecture of ADBL2, represented in Figure \ref{fig:architecture}, can be divided in two main parts. 
\begin{enumerate}
    \item The Web UI which consists in a web application where the user can import an argumentation tree (using Kialo's format), explore it, apply changes, and export the result argumentation tree.

    \item The inference core of ADBL2 translates the user input according to the prompt engineering technique (e.g., adding a few-shot priming or not) and the LLM chosen by the user (different LLMs have different prompt templates) into a final prompt. This inference core performs RBAM: the output of the LLM is constrained using LMQL\footnote{\url{https://lmql.ai/}} to obtain the probability to predict each label ("attack" and "support") which is given to the user via the Web UI.

\end{enumerate}





\section{A Fine-tuned LLM for relation-based mining}
\label{sec:fine-tined}

\subsection{Datasets}
\label{sec:dataset}
Our test dataset $\D$ consists of triples $(x,y,z) \in \D$ such that $(x,y)$ is a pair of argument and $z\in \{attack, support\}$ is the type of the relation from $x$ to $y$. We collected these triples by exporting debates on various domains (Art, Climate Change, etc.) from Kialo between the 8th and 15th of March 2024. 
We made use of the random undersampling algorithm from the imbalanced-learn library\footnote{\url{https://imbalanced-learn.org/}}, first by domain and by relation type, to obtain a balanced dataset. 
The number of triples per domain is displayed in Table \ref{table-eval}.

While it is not possible to reproduce the baseline protocol of Gorur et al. \cite{DBLP:journals/corr/abs-2402-11243} (as they do not provide the Kialo dataset they used), we wanted to get as close as possible to their settings.
We created a similar dataset $\D_{l,p,s}$ of arguments related to law, politics and sports debates. This dataset was separated in a train ($\D_{l,p,s}^{\mathtt{Train}}$ with $\D_{l,p,s}^{\mathtt{Train}} \cap \D = \emptyset$) and test ($\D_{l,p,s}^{\mathtt{Test}} \subseteq \D)$  datasets, with a 77.8/22.2 split, while preserving class balance. 

Given a Kialo bipolar argumentation tree $\AS= (\A,\S,\C, r)$, where $\A$ is a set of arguments, $\S \subseteq \A \times \A$ is a binary support relation between arguments, and $\C \subseteq \A \times \A$ is a binary attack relation, and $r$ is the root of the tree, the depth of an argument $a \in \A$ is $n$ iff there exists a sequence of arguments $(a_0, a_1, \dots, a_n)$ with $a_n = r$, $a_0 = a$, and $(a_i, a_{i+1}) \in \C \cup \S$ for all $0 \leq i \leq n-1.$
Note that to ensure a high quality dataset for the training of our language model $(\D_{l,p,s})$, we only extracted the pair of arguments closer to the root as they were more explored by the Kialo community and thus more refined. 
Namely, we only extracted the triples $(x,y,z)$ such that the depth of $x$ is less or equal to $7$.

\subsection{Fine-tuning Mistral}
\label{sec:fine-tuned}

For the fine-tuning, we used a Linux virtual machine with a 12-core Intel Xeon Processor (Skylake, IBRS), 125 Gb of RAM, and a NVIDIA A40 with 46Gb of VRAM.
Our main goal was to restrict ourselves to large language models that can be run on consumer hardware.
We selected Mistral-7B \cite{jiang2023mistral} as our LLM as it was best performing LLM that could be run and fine-tuned on our setting.

Since a full fine-tuning of the model was not possible, we used a Parameter-Efficient Fine-Tuning technique (PEFT) called Low Rank Adaptation (LoRA) \cite{DBLP:conf/iclr/HuSWALWWC22} which reduces the VRAM consumption during the fine-tuning process. Namely, additional parameters are added to the model, and only those are trained while the initial parameters of the large language model are frozen.
We also used QLoRA \cite{DBLP:conf/nips/DettmersPHZ23} to further reduced the VRAM consumption, i.e., the LLM parameters are quantised to 8 bits (instead of 16 bits) before the fine-tuning.

\begin{figure}[!h]
    \centering
    \includegraphics[width=5.5cm]{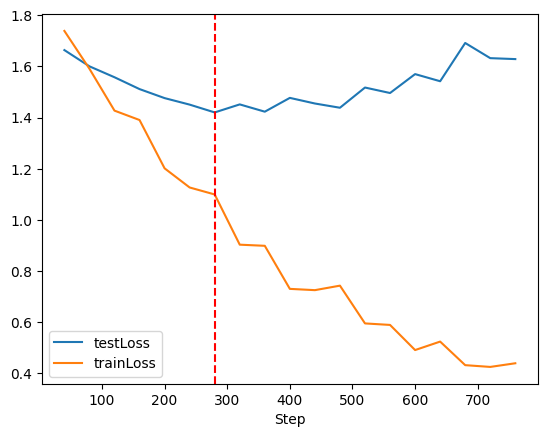}
    \caption{Plot of the loss ($y$-axis) on the training (orange) and test (blue) datasets during the fine-tuning per training iteration ($x$-axis).}
    \label{fig:train-loss}
\end{figure}

Mistral 7B was fine-tuned on $\D_{l,p,s}^{\mathtt{Train}}$, the training dataset  of $\D_{l,p,s}$. 
Each triple $(x,y,z) \in \D_{l,p,s}^{\mathtt{Train}}$ was transformed into a prompt using $x$ and $y$ (see the prompt in Figure \ref{fig:architecture}). With this prompt as input, the LLM must predict a token $\hat{z} \in \{attack, support\}$ which must correspond to $z$.
The training parameters are $r=8$, $\mathtt{lora\_alpha}=16$, $\mathtt{lora\_dropout}=0.1$, $\mathtt{per\_device\_train\_batch\_size}=16$, $\mathtt{learning\_rate}=1e-4$, and $\mathtt{bias}=None$.
We used an early stopping approach with a monitor on the loss. The final fine-tuned model was trained for 280 training steps (see Figure \ref{fig:train-loss}).

The fine-tuned model is available at: \url{https://huggingface.co/4mbroise/ADBL2-Mistral-7B}.

{\tiny
\begin{table*}[!h]
\centering
\begin{tabular}{|c|c|c|c|c|}
\cline{2-5}
                    \multicolumn{1}{c}{} & \multicolumn{2}{|c|}{Test data $\D$}            & Mistral 7B-16bits + 4-Shots        & Fine-tuned Mistral 7B            \\
\cline{2-5}
                    \multicolumn{1}{c|}{}  & \textit{Attack} & \textit{Support} & \textit{Attack/Support/Macro} F1-score & \textit{Attack/Support/Macro} F1-score  \\
\hline
Art                  & 94              & 129              & 73.1 / 83.9 / 78.5           & \textbf{89.5} / \textbf{92.1} / \textbf{90.8   }        \\
Climate Change       & 419             & 508              & 66.6 / 82.1 / 74.3           &\textbf{ 93.3} / \textbf{94.5 } / \textbf{93.9}           \\
Economics            & 298             & 298              & 72.0 / 79.8 / 75.9           & \textbf{90.0} / \textbf{90.1} / \textbf{90.3}           \\
Entertainment        & 490             & 612              & 64.3 / 81.9 / 73.1           & \textbf{92.0} / \textbf{93.5} / \textbf{92.7}           \\
Health               & 355             & 473              & 64.5 / 81.7 / 73.1           & \textbf{90.8} / \textbf{93.3} / \textbf{92.2 }          \\
Lgbtq                & 277             & 338              & 67.4 / 80.9 / 74.2           & \textbf{90.9} / \textbf{92.4} / \textbf{ 91.6  }         \\
Life                 & 353             & 352              & 81.5 / 84.2 / 82.9           & \textbf{90.8} / \textbf{90.5} / \textbf{90.6}           \\
Privacy              & 164             & 167              & 71.5 / 79.9 / 75.7           & \textbf{89.7} / \textbf{89.8} / \textbf{ 89.7 }          \\
Law, Politics, Sports & 891             & 867             & 69.2 / 78.8 / 74.0           & \textbf{91.9} / \textbf{91.8} / \textbf{91.8}           \\
Technology           & 537             & 554              & 67.2 / 79.2 / 73.2           &\textbf{ 92.0 } / \textbf{92.6} / \textbf{92.3}          \\
\hline
\end{tabular}
\label{table-eval}
\caption{Evaluation of Mistral 7B-16bits with few-shot priming and our fine-tuned Mistral 7B models on our test dataset.}
\end{table*}}

\subsection{Evaluation}

We evaluated the performance and generalisation capabilities of our new quantised fine-tuned Mistral 7B model (as described in Section \ref{sec:fine-tuned}).
As a baseline, we use Mistral 7B-16bit\footnote{\url{https://huggingface.co/mistralai/Mistral-7B-v0.1}} with a few-shot priming composed of the same four fixed pair of argument examples, similar to \cite{DBLP:journals/corr/abs-2402-11243}. 
To constrain the output generated by the two LLMs to $\{attack, support\}$, we used LMQL as described in Section \ref{sec:tool}.

In Table \ref{table-eval}, we reported the attack (resp. support) F1-score of the two LLMs as well as the macro F1-score.
We can see that our new fine-tuned model outperforms the Mistral 7B-16bit model equipped with the few-shot priming on all domains. Moreover, we can see that while we only fine-tuned our LLM on the law, politics, and sports domains, the model performance on all domains increased significantly, achieving an average macro F1-score of 90.59$\%$ across all domains.


\section{Discussion and Future Work}
\label{sec:future_work}


In this paper, we introduced ADBL2, an assisted debate builder tool. 
It is based on the capability of large language models to generalise and perform relation-based argument mining in a wide-variety of domains.
It is the first open-source tool that leverages relation-based mining for (1) the verification of existing relations in a debate and (2) the assisted creation of new arguments by means of large language models.
ADBL2 is highly modular and can work with any open-source large language models that are used as plugins.
As a by-product, we also provide the first fine-tuned Mistral-7B large language model for relation-based argument mining, usable by ADBL2, which outperforms existing approaches for this task with an overall F1-score of 90.59\% across all domains.

While this work shows promising results for RBAM, we still need to assess the generalisation capabilities of our fine-tuned Mistral 7B model on other argumentative datasets (e.g, Essays, Nixon-Kennedy, etc.). Moreover, we would also need to extend the model to perform ternary RBAM to identify arguments that are not related. We also plan to explore other types of LLMs such as heavily quantised models, pruned LLMs \cite{DBLP:conf/nips/MaFW23}, more recent LLMs (e.g., Llama 3\footnote{\url{https://llama.meta.com/llama3/}}, Gemma \cite{gemmateam2024gemma}), or LLMs fine-tuned with other PEFT techniques \cite{DBLP:journals/corr/abs-2403-03507,DBLP:conf/nips/LiuTMMHBR22,lialin2023relora}.

\noindent {\bf Ethics statement} We note that while there are risks to LLMs such as bias and misinformation, we only use LLMs to generate a single token, which is support/attack. Thus, there are no risks of generating biased or false information.

\bibliographystyle{unsrt}  
\bibliography{templateArxiv}

\begin{thebibliography}{10}

\bibitem{DBLP:conf/sum/CroitoruV13}
Madalina Croitoru and Srdjan Vesic.
\newblock What can argumentation do for inconsistent ontology query answering?
\newblock In Weiru Liu, V.~S. Subrahmanian, and Jef Wijsen, editors, {\em
  Scalable Uncertainty Management - 7th International Conference, {SUM} 2013,
  Washington, DC, USA, September 16-18, 2013. Proceedings}, volume 8078 of {\em
  Lecture Notes in Computer Science}, pages 15--29. Springer, 2013.

\bibitem{DBLP:phd/hal/Yun19}
Bruno Yun.
\newblock {\em Argumentation techniques for existential rules. (Techniques
  d'argumentation pour les r{\`{e}}gles existentielles)}.
\newblock PhD thesis, University of Montpellier, France, 2019.

\bibitem{DBLP:journals/ai/Dung95}
Phan~Minh Dung.
\newblock On the acceptability of arguments and its fundamental role in
  nonmonotonic reasoning, logic programming and n-person games.
\newblock {\em Artif. Intell.}, 77(2):321--358, 1995.

\bibitem{DBLP:journals/ker/BaroniCG11}
Pietro Baroni, Martin Caminada, and Massimiliano Giacomin.
\newblock An introduction to argumentation semantics.
\newblock {\em Knowl. Eng. Rev.}, 26(4):365--410, 2011.

\bibitem{DBLP:conf/comma/Caminada06}
Martin Caminada.
\newblock Semi-stable semantics.
\newblock In Paul~E. Dunne and Trevor J.~M. Bench{-}Capon, editors, {\em
  Computational Models of Argument: Proceedings of {COMMA} 2006, September
  11-12, 2006, Liverpool, {UK}}, volume 144 of {\em Frontiers in Artificial
  Intelligence and Applications}, pages 121--130. {IOS} Press, 2006.

\bibitem{DBLP:conf/sum/AmgoudB13a}
Leila Amgoud and Jonathan Ben{-}Naim.
\newblock Ranking-based semantics for argumentation frameworks.
\newblock In Weiru Liu, V.~S. Subrahmanian, and Jef Wijsen, editors, {\em
  Scalable Uncertainty Management - 7th International Conference, {SUM} 2013,
  Washington, DC, USA, September 16-18, 2013. Proceedings}, volume 8078 of {\em
  Lecture Notes in Computer Science}, pages 134--147. Springer, 2013.

\bibitem{DBLP:conf/aaai/BonzonDKM16}
Elise Bonzon, J{\'{e}}r{\^{o}}me Delobelle, S{\'{e}}bastien Konieczny, and
  Nicolas Maudet.
\newblock A comparative study of ranking-based semantics for abstract
  argumentation.
\newblock In Dale Schuurmans and Michael~P. Wellman, editors, {\em Proceedings
  of the Thirtieth {AAAI} Conference on Artificial Intelligence, February
  12-17, 2016, Phoenix, Arizona, {USA}}, pages 914--920. {AAAI} Press, 2016.

\bibitem{DBLP:conf/comma/YunVCB18}
Bruno Yun, Srdjan Vesic, Madalina Croitoru, and Pierre Bisquert.
\newblock Viewpoints using ranking-based argumentation semantics.
\newblock In Sanjay Modgil, Katarzyna Budzynska, and John Lawrence, editors,
  {\em Computational Models of Argument - Proceedings of {COMMA} 2018, Warsaw,
  Poland, 12-14 September 2018}, volume 305 of {\em Frontiers in Artificial
  Intelligence and Applications}, pages 381--392. {IOS} Press, 2018.

\bibitem{DBLP:conf/nmr/AmgoudCL04}
Leila Amgoud, Claudette Cayrol, and Marie{-}Christine Lagasquie{-}Schiex.
\newblock On the bipolarity in argumentation frameworks.
\newblock In James~P. Delgrande and Torsten Schaub, editors, {\em 10th
  International Workshop on Non-Monotonic Reasoning {(NMR} 2004), Whistler,
  Canada, June 6-8, 2004, Proceedings}, pages 1--9, 2004.

\bibitem{DBLP:conf/ecsqaru/CayrolL05}
Claudette Cayrol and Marie{-}Christine Lagasquie{-}Schiex.
\newblock Gradual valuation for bipolar argumentation frameworks.
\newblock In Llu{\'{\i}}s Godo, editor, {\em Symbolic and Quantitative
  Approaches to Reasoning with Uncertainty, 8th European Conference, {ECSQARU}
  2005, Barcelona, Spain, July 6-8, 2005, Proceedings}, volume 3571 of {\em
  Lecture Notes in Computer Science}, pages 366--377. Springer, 2005.

\bibitem{DBLP:conf/sgai/HimeurYBC21}
Areski Himeur, Bruno Yun, Pierre Bisquert, and Madalina Croitoru.
\newblock Assessing the impact of agents in weighted bipolar argumentation
  frameworks.
\newblock In Max Bramer and Richard Ellis, editors, {\em {SGAI} 2021,
  Proceedings}, volume 13101 of {\em Lecture Notes in Computer Science}, pages
  75--88. Springer, 2021.

\bibitem{DBLP:conf/argmas/NielsenP06}
S{\o}ren~Holbech Nielsen and Simon Parsons.
\newblock A generalization of dung's abstract framework for argumentation:
  Arguing with sets of attacking arguments.
\newblock In Nicolas Maudet, Simon Parsons, and Iyad Rahwan, editors, {\em
  Argumentation in Multi-Agent Systems, Third International Workshop, ArgMAS
  2006, Hakodate, Japan, May 8, 2006, Revised Selected and Invited Papers},
  volume 4766 of {\em Lecture Notes in Computer Science}, pages 54--73.
  Springer, 2006.

\bibitem{DBLP:conf/aaai/YunVC20}
Bruno Yun, Srdjan Vesic, and Madalina Croitoru.
\newblock Ranking-based semantics for sets of attacking arguments.
\newblock In {\em The Thirty-Fourth {AAAI} Conference on Artificial
  Intelligence, {AAAI} 2020, The Thirty-Second Innovative Applications of
  Artificial Intelligence Conference, {IAAI} 2020, The Tenth {AAAI} Symposium
  on Educational Advances in Artificial Intelligence, {EAAI} 2020, New York,
  NY, USA, February 7-12, 2020}, pages 3033--3040. {AAAI} Press, 2020.

\bibitem{DBLP:journals/jair/HunterT17}
Anthony Hunter and Matthias Thimm.
\newblock Probabilistic reasoning with abstract argumentation frameworks.
\newblock {\em J. Artif. Intell. Res.}, 59:565--611, 2017.

\bibitem{DBLP:conf/jelia/CramerG19}
Marcos Cramer and Mathieu Guillaume.
\newblock Empirical study on human evaluation of complex argumentation
  frameworks.
\newblock In Francesco Calimeri, Nicola Leone, and Marco Manna, editors, {\em
  Logics in Artificial Intelligence - 16th European Conference, {JELIA} 2019,
  Rende, Italy, May 7-11, 2019, Proceedings}, volume 11468 of {\em Lecture
  Notes in Computer Science}, pages 102--115. Springer, 2019.

\bibitem{DBLP:conf/atal/VesicYT22}
Srdjan Vesic, Bruno Yun, and Predrag Teovanovic.
\newblock Graphical representation enhances human compliance with principles
  for graded argumentation semantics.
\newblock In Piotr Faliszewski, Viviana Mascardi, Catherine Pelachaud, and
  Matthew~E. Taylor, editors, {\em 21st International Conference on Autonomous
  Agents and Multiagent Systems, {AAMAS} 2022, Auckland, New Zealand, May 9-13,
  2022}, pages 1319--1327. International Foundation for Autonomous Agents and
  Multiagent Systems {(IFAAMAS)}, 2022.

\bibitem{DBLP:conf/ijcai/CramerG18}
Marcos Cramer and Mathieu Guillaume.
\newblock Directionality of attacks in natural language argumentation.
\newblock In Claudia Schon, editor, {\em Proceedings of the fourth Workshop on
  Bridging the Gap between Human and Automated Reasoning {(IJCAI-ECAI} 2018),
  Stockholm, Schweden, July 14, 2018}, volume 2261 of {\em {CEUR} Workshop
  Proceedings}, pages 40--46. CEUR-WS.org, 2018.

\bibitem{DBLP:conf/naacl/CarstensT15}
Lucas Carstens and Francesca Toni.
\newblock Towards relation based argumentation mining.
\newblock In {\em Proceedings of the 2nd Workshop on Argumentation Mining,
  ArgMining@HLT-NAACL 2015, June 4, 2015, Denver, Colorado, {USA}}, pages
  29--34. The Association for Computational Linguistics, 2015.

\bibitem{DBLP:conf/lrec/KonatLPBR16}
Barbara Konat, John Lawrence, Joonsuk Park, Katarzyna Budzynska, and Chris
  Reed.
\newblock A corpus of argument networks: Using graph properties to analyse
  divisive issues.
\newblock In Nicoletta Calzolari, Khalid Choukri, Thierry Declerck, Sara Goggi,
  Marko Grobelnik, Bente Maegaard, Joseph Mariani, H{\'{e}}l{\`{e}}ne Mazo,
  Asunci{\'{o}}n Moreno, Jan Odijk, and Stelios Piperidis, editors, {\em
  Proceedings of the Tenth International Conference on Language Resources and
  Evaluation {LREC} 2016, Portoro{\v{z}}, Slovenia, May 23-28, 2016}. European
  Language Resources Association {(ELRA)}, 2016.

\bibitem{DBLP:journals/coling/LawrenceR19}
John Lawrence and Chris Reed.
\newblock Argument mining: {A} survey.
\newblock {\em Comput. Linguistics}, 45(4):765--818, 2019.

\bibitem{DBLP:journals/artmed/MayerMCV21}
Tobias Mayer, Santiago Marro, Elena Cabrio, and Serena Villata.
\newblock Enhancing evidence-based medicine with natural language argumentative
  analysis of clinical trials.
\newblock {\em Artif. Intell. Medicine}, 118:102098, 2021.

\bibitem{DBLP:journals/expert/Ruiz-DolzABG21}
Ramon Ruiz{-}Dolz, Jos{\'{e}} Alemany, Stella~Heras Barber{\'{a}}, and Ana
  Garc{\'{\i}}a{-}Fornes.
\newblock Transformer-based models for automatic identification of argument
  relations: {A} cross-domain evaluation.
\newblock {\em {IEEE} Intell. Syst.}, 36(6):62--70, 2021.

\bibitem{DBLP:journals/corr/abs-2402-11243}
Deniz Gorur, Antonio Rago, and Francesca Toni.
\newblock Can large language models perform relation-based argument mining?
\newblock {\em CoRR}, abs/2402.11243, 2024.

\bibitem{touvron2023llama}
Hugo Touvron, Louis Martin, Kevin Stone, Peter Albert, Amjad Almahairi, Yasmine
  Babaei, Nikolay Bashlykov, Soumya Batra, Prajjwal Bhargava, Shruti Bhosale,
  et~al.
\newblock Llama 2: Open foundation and fine-tuned chat models.
\newblock {\em arXiv preprint arXiv:2307.09288}, 2023.

\bibitem{jiang2023mistral}
Albert~Q Jiang, Alexandre Sablayrolles, Arthur Mensch, Chris Bamford,
  Devendra~Singh Chaplot, Diego de~las Casas, Florian Bressand, Gianna Lengyel,
  Guillaume Lample, Lucile Saulnier, et~al.
\newblock Mistral 7b.
\newblock {\em arXiv preprint arXiv:2310.06825}, 2023.

\bibitem{DBLP:conf/iclr/HuSWALWWC22}
Edward~J. Hu, Yelong Shen, Phillip Wallis, Zeyuan Allen{-}Zhu, Yuanzhi Li,
  Shean Wang, Lu~Wang, and Weizhu Chen.
\newblock Lora: Low-rank adaptation of large language models.
\newblock In {\em The Tenth International Conference on Learning
  Representations, {ICLR} 2022, Virtual Event, April 25-29, 2022}.
  OpenReview.net, 2022.

\bibitem{DBLP:conf/nips/DettmersPHZ23}
Tim Dettmers, Artidoro Pagnoni, Ari Holtzman, and Luke Zettlemoyer.
\newblock Qlora: Efficient finetuning of quantized llms.
\newblock In Alice Oh, Tristan Naumann, Amir Globerson, Kate Saenko, Moritz
  Hardt, and Sergey Levine, editors, {\em Advances in Neural Information
  Processing Systems 36: Annual Conference on Neural Information Processing
  Systems 2023, NeurIPS 2023, New Orleans, LA, USA, December 10 - 16, 2023},
  2023.

\bibitem{DBLP:conf/nips/MaFW23}
Xinyin Ma, Gongfan Fang, and Xinchao Wang.
\newblock Llm-pruner: On the structural pruning of large language models.
\newblock In Alice Oh, Tristan Naumann, Amir Globerson, Kate Saenko, Moritz
  Hardt, and Sergey Levine, editors, {\em Advances in Neural Information
  Processing Systems 36: Annual Conference on Neural Information Processing
  Systems 2023, NeurIPS 2023, New Orleans, LA, USA, December 10 - 16, 2023},
  2023.

\bibitem{gemmateam2024gemma}
Gemma~Team et~al.
\newblock Gemma: Open models based on gemini research and technology, 2024.

\bibitem{DBLP:journals/corr/abs-2403-03507}
Jiawei Zhao, Zhenyu Zhang, Beidi Chen, Zhangyang Wang, Anima Anandkumar, and
  Yuandong Tian.
\newblock Galore: Memory-efficient {LLM} training by gradient low-rank
  projection.
\newblock {\em CoRR}, abs/2403.03507, 2024.

\bibitem{DBLP:conf/nips/LiuTMMHBR22}
Haokun Liu, Derek Tam, Mohammed Muqeeth, Jay Mohta, Tenghao Huang, Mohit
  Bansal, and Colin Raffel.
\newblock Few-shot parameter-efficient fine-tuning is better and cheaper than
  in-context learning.
\newblock In Sanmi Koyejo, S.~Mohamed, A.~Agarwal, Danielle Belgrave, K.~Cho,
  and A.~Oh, editors, {\em Advances in Neural Information Processing Systems
  35: Annual Conference on Neural Information Processing Systems 2022, NeurIPS
  2022, New Orleans, LA, USA, November 28 - December 9, 2022}, 2022.

\bibitem{lialin2023relora}
Vladislav Lialin, Namrata Shivagunde, Sherin Muckatira, and Anna Rumshisky.
\newblock Relora: High-rank training through low-rank updates, 2023.

\end{thebibliography}

\end{document}